\documentclass[runningheads]{llncs}
\usepackage[T1]{fontenc}
\usepackage{booktabs}
\usepackage{newtxtext}       % 
\usepackage[varvw]{newtxmath}       % selects Times Roman as basic font

\usepackage{graphicx}
\begin{document}
\title{AutoAdapt: Automatic Domain Discovery Enables Low-Cost Extensibility}
%
% \titlerunning{AutoAdapt: Automatic Domain Specialisation}
\author{Josh McGiff\orcidID{0009-0001-8209-3348} \and
Salma Mekaoui\orcidID{0009-0005-2826-060X} \and
Robert Shanahan\orcidID{0009-0001-6725-6954} \and 
Nikola S. Nikolov\orcidID{0000-0001-8022-0297}}
\authorrunning{J. McGiff et al.}
\institute{University of Limerick, Ireland\\
\email{mcgiff.josh@ul.ie}}
\maketitle              % typeset the header of the contribution
\begin{abstract}
Instruction-tuned models are deployed into environments where domains are heterogeneous and evolve, yet adding new domains or data typically requires costly retraining. We present AutoAdapt, a modular framework that incorporates new domains and data via targeted single-adapter training without modifying other adapters. The framework automatically discovers latent domains, uses them to train per-domain Low-Rank Adaptation (LoRA) adapters independently in parallel and performs parameter-free routing. Across 14 domain-specific benchmarks and GPT-4o pairwise judgements, AutoAdapt achieves parity with a LoRA adapter trained on all domains without requiring full-model retraining. We also find evidence of specialisation effect convergence across independent discovery methods. Overall, training each adapter on its own domain prevents domain interference by construction, thus enabling modular, taxonomy-free domain specialisation without aggregate performance loss or full model retraining.  

\end{abstract}
\section{Introduction}
Supervised instruction fine-tuning has become the 
standard approach for adapting large language models 
to follow human instructions~\cite{wei2021finetuned,ouyang2022training}, yet the field has optimised for 
benchmark performance while leaving deployment 
maintainability largely 
unsolved~\cite{kiela-etal-2021-dynabench}. Models 
are evaluated on static benchmarks but deployed into 
environments where domains evolve and new 
capabilities are continuously required. However, full retraining on 
ever-growing combined datasets poses a structural barrier 
for practitioners without access to large-scale 
compute~\cite{mcgiff2025overcomingdatascarcitygenerative}.

Although parameter-efficient fine-tuning methods such as Low-Rank Adaptation (LoRA) reduce the computational cost of instruction tuning \cite{hu2021loralowrankadaptationlarge}, a single adapter trained on heterogeneous data inherits two problems. First, heterogeneous training data introduces 
conflicting gradient signals across 
domains~\cite{NEURIPS2020_3fe78a8a,ling2025domain}, 
which limits the degree to which a single model 
can specialise~\cite{NEURIPS2023_ec641387}. Second, monolithic 
fine-tuning produces brittle, difficult-to-extend systems \cite{Huang_Yang_Wang_Qi_Yu_Fan_Wang_2026}. In such cases, incorporating new domains or updating existing knowledge
requires full retraining on the combined dataset, risking 
catastrophic forgetting of previously learned behaviours 
\cite{MCCLOSKEY1989109,li-etal-2024-revisiting}. 

Standard Mixture-of-Experts (MoE) architectures address 
specialisation through jointly trained experts and 
routing components~\cite{shazeer2017outrageouslylargeneuralnetworks}, 
but do not resolve the extensibility problem. Joint 
training couples all experts together, meaning 
new domains cannot be added without retraining 
the full system. A study \cite{NEURIPS2023_ec641387} 
confirms that instruction-tuned models on 
heterogeneous data fail to achieve consistent 
cross-domain performance, identifying modularity 
as a key unmet need. Although prior work~\cite{mcgiff2025semiadaptsemiloraefficientdomain} demonstrates that 
fixed-domain specialisation enables parameter-efficient 
methods to match full-model 
fine-tuning baselines in low-resource translation, the study does not provide empirical evidence of extensibility.

In this work, we introduce AutoAdapt, a modular framework that eliminates the retraining bottleneck for heterogeneous instruction tuning. Rather than assuming a fixed domain taxonomy, AutoAdapt automatically discovers latent domains by comparing unsupervised clustering (K-means, BERTopic) and zero-shot classification (MNLI) methods. Domain adapters are trained independently and in parallel with parameter-free routing occurring at inference via cosine similarity to domain centroids. AutoAdapt enables targeted domain adapter creation and updates, thus circumventing the need for full-model retraining. We contribute: a \textbf{systematic cross-method comparison} of three automatic domain discovery methods for instruction tuning; \textbf{AutoAdapt},\textbf{ a} \textbf{modular instruction tuning framework} that matches single LoRA 
    adapter performance; and a post-deployment \textbf{extensibility demonstration}. 

% \begin{itemize}
    
%     \item \textbf{Empirical Study.} .
        
%     \item \textbf{System.} AutoAdapt, a modular 
%     instruction tuning framework that matches single LoRA 
%     adapter performance, while enabling domain updates that monolithic approaches 
%     cannot support.
%     \item \textbf{Extensibility Demonstration.} 
%     Post-deployment domain update simulations  
%     showing that domain extensibility is a structural property of AutoAdapt.
%     % extensibility analysis mby instead if i get the extension results
% \end{itemize}

\section{Related Work}
\subsection{Instruction Tuning and Data Heterogeneity}

Instruction tuning has emerged as a fundamental post-training step for aligning large language models with human intent and expectations \cite{wei2021finetuned,ouyang2022training}. However, ever-expanding instruction datasets contribute to inherent heterogeneity. This presents a central challenge where training on dissimilar tasks introduces conflicting gradient signals and consequently degrades individual task performance \cite{NEURIPS2023_ec641387}.

Recent work addresses instruction data 
efficiency through filtering and quality 
valuation methods rather than structural 
partitioning. Centroid-based clustering with confidence-guided selection has been used to curate high-quality subsets for efficient training of instruction tuning models \cite{cai-etal-2025-low}. Similarly, ROSE~\cite{wu-etal-2025-rose} leverages 
pairwise preference loss as a reward signal 
to select high-utility samples for 
task-specific instruction tuning, while LimaCost~\cite{moon-etal-2025-limacost} 
estimates sample value by measuring 
gradient proximity to the LIMA dataset, 
selecting data that maximises alignment 
performance. These approaches 
improve data efficiency but discard low-scored 
samples, risking reduced generalisation to 
out-of-distribution examples. AutoAdapt 
takes an alternative approach, partitioning the 
full instruction dataset into domain-specific 
subsets without discarding any examples, with 
each assigned to exactly one domain adapter.

   \subsection{Parameter-Efficient Fine-Tuning and Adapter Routing}
Parameter-efficient fine-tuning (PEFT) methods reduce the cost of adapting large pre-trained models by updating only a small fraction of parameters. One such method, Low-Rank Adaptation  \cite{hu2021loralowrankadaptationlarge}, achieves competitive performance with full fine-tuning at significantly reduced memory cost by expressing weight updates as low-rank matrix decompositions. However, catastrophic forgetting is a well-documented challenge in heterogeneous instruction tuning settings, whereby updating model parameters on a new task or domain degrades prior capabilities \cite{MCCLOSKEY1989109,li-etal-2024-revisiting}. Recent work addresses this challenge through constrained LoRA updates. CLoRA~\cite{lu-etal-2025-controlled} introduces 
orthogonal subspace regularisation to preserve 
general capabilities during sequential 
fine-tuning of large language models. Alternatively, SLIM \cite{han-etal-2025-slim} proposes dynamic routing between LoRA adapters and identity layers to suppress forgetting. Pletenev~\emph{et al.}~\cite{pletenev-etal-2025-much}
demonstrate that incorporating new knowledge 
via LoRA risks degrading previously learned 
capabilities, with bias amplified when training 
data is skewed towards specific entities.

The proliferation of domain-specific adapters has motivated work on multi-adapter routing. Tian ~\emph{et al.} \cite{tian-etal-2025-adapters} propose training a dedicated selector adapter to route between domain and task-specific LoRA modules at inference time. Similarly, GLIDER~\cite{li-etal-2025-glider} constructs 
LLM-generated global routing vectors combined 
with local task vectors to select from a pool 
of PEFT experts. Unlike these approaches, which require either a trained routing component or LLM-generated routing signals, our approach extends parameter-free routing to the instruction tuning setting.

    % OG SemiAdapt did peft for domains
    % are there other instruction tuning papers that did peft 
    % any instruction tuning with peft and for domains
    % Adapter routing 
    \vspace{-0.5em}

% \subsection{Domain Discovery Methods}
%     Unsupervised discovery of latent structure in text 
% has a long history in NLP. Latent Dirichlet 
% Allocation~\cite{blei2003latent} introduced 
% probabilistic topic modelling, representing documents 
% as mixtures of topics and topics as distributions 
% over words. Neural approaches such as 
% BERTopic~\cite{grootendorst2022bertopic} and 
% Top2Vec~\cite{angelov2020top2vec} extend this by 
% clustering dense sentence embeddings, producing 
% more semantically coherent topics. K-Means clustering has been used to identify representative subsets of full instruction datasets \cite{yu2024diversifyconquerdiversitycentricdata}. Zero-shot classification via natural language 
% inference enables domain assignment using 
% predefined label categories without task-specific 
% training~\cite{yin-etal-2019-benchmarking}. AutoAdapt 
% conducts, to our knowledge, the first systematic comparison of clustering, topic modelling, 
% and zero-shot classification, under identical 
% controlled conditions for instruction tuning 
% domain discovery.

\section{Methodology}
We discover latent domains in heterogeneous instruction data using three independent methods, train a dedicated LoRA adapter per domain, and route for inference with a parameter-free centroid-based mechanism.

\subsection{Latent Domain Discovery}
We discover latent domains via three independent methods. For unsupervised clustering, we embed each training instruction with all-MiniLM-L6-v2~\cite{NEURIPS2020_3f5ee243} and apply K-means, with $K$ selected via Mini-Batch K-means elbow analysis and confirmed via silhouette/Calinski-Harabasz scores, converging on $K=17$. Clustering is then run on the complete dataset at $K=17$. For topic modelling, we use BERTopic~\cite{grootendorst2022bertopic}, which applies contextual-embedding-based clustering and automatically determines domain count~\cite{robledo2022topic}, yielding 107 topics, which we additionally merge via intertopic distance into a coarser 24-domain super-grouping. For zero-shot classification, we apply \texttt{facebook/bart-large-mnli}, entailing each instruction against hypotheses of the form \textit{"This text is about [label]"} using the same 24-label space as the BERTopic super-grouping, which reflects a realistic deployment scenario where an existing taxonomy labels new data without retraining.

\subsection{Adapter Training and Centroid-Based Routing}
\label{sec:routing}
We train a dedicated LoRA adapter~\cite{hu2021loralowrankadaptationlarge} per discovered domain and discovery method, with the base model frozen throughout. Additionally, domain discovery, adapter training, and routing are fully decoupled such that new domains only require one additional adapter without needing to retrain other components. We compare against a single LoRA adapter trained on all domains. At inference, we route test instructions using a parameter-free, centroid-based mechanism: for each domain $k$ we precompute a centroid $\mathbf{c}_k$ as the mean embedding of its training instructions,
\begin{equation}
    \mathbf{c}_k = \frac{1}{|D_k|} \sum_{x_i \in D_k} \phi(x_i),
\end{equation}
and route a test instruction $x$ to the domain with highest cosine similarity,
\begin{equation}
    \hat{k} = \arg\max_{k} \frac{\phi(x) \cdot \mathbf{c}_k}{\|\phi(x)\| \, \|\mathbf{c}_k\|}.
\end{equation}
This adds no trainable parameters and requires a single embedding forward pass per inference call.

\subsection{Training Data and Evaluation}
We train the Mistral-7B-Instruct-v0.3 base model on 509,174 instructions drawn from eight public datasets (shown in Table \ref{table:dataset}) spanning general instruction-following, code, QA, finance, medicine, and reasoning. We compare the performance of AutoAdapt and single LoRA on domain-specific benchmarks and blind pairwise GPT-4o judge evaluation~\cite{NEURIPS2023_91f18a12}. 

\begin{table}[htbp]
\centering
\tiny
\caption{Composition and statistics of the instruction-tuning dataset.}
\begin{tabular}{lccc}
\hline
\textbf{Dataset} & \textbf{Domain} & 
\textbf{\#Samples} & \textbf{Avg Tokens} \\
\hline
Alpaca-cleaned          & General      & 51,760  & $\sim$120 \\
CodeAlpaca-20K          & Code         & 20,022  & $\sim$40  \\
WebGPT                  & QA/Retrieval & 18,994  & $\sim$169 \\
Databricks Dolly 15K    & General      & 15,011  & $\sim$129 \\
Finance-Alpaca          & Finance      & 68,912  & $\sim$162 \\
AlpaCare-MedInstruct    & Medicine     & 52,002  & $\sim$184 \\
FLAN (CoT)              & Reasoning    & 74,771  & $\sim$70  \\
StackOverflow Dialogues & Programming  & 207,702 & $\sim$130 \\
\hline
\textbf{Total}          &              & \textbf{509,174} & \\
\hline
\end{tabular}
\label{table:dataset}
\end{table}

\section{Results and Discussion}

\subsection{Cross-Method Convergence}
% section on significance adapt below for it
% first report the gpt-4o results, then the benchmarks
In this section we explore a mixture of LLM-as-a-judge and domain-specific evaluation for our three independent automatically discovered latent domain specialisation approaches.  
\subsubsection{LLM-as-a-Judge Evaluation}
AutoAdapt achieves performance comparable with the single-LoRA baseline on LLM-as-a-judge evaluation. For each domain, we sample 100 instances from the test set and compare the output of the corresponding domain-specific adapter against that of the single LoRA model. Comparisons are conducted with GPT-4o in a blind, reference-grounded pairwise preference setting. Adapter win rates against the single LoRA were 50.5\% for K-means (858/1,700), 50.2\% for BERTopic (1,205/2,400), and 46.9\% for MNLI (1,004/2,143).

% heatmap across all 3 methods required here

% \begin{table}[htbp]
% \centering
% \caption{LLM-as-a-judge evaluation results per AutoAdapt method using GPT-4o in a blind, reference-grounded pairwise preference scenario. Up to N=100 samples are selected per domain  and compared with the corresponding domain adapter and single LoRA inference output. Winrate is the ratio of AutoAdapt method wins over the total number of comparisons.}
% \tiny
% \begin{tabular}{llrr}
% \toprule
% \textbf{Method} & \textbf{\# of domains} & \textbf{Centroid Wins} & \textbf{Winrate} \\
% \midrule
% K-means & 17 & 858/1700 & 50.5 \\
% BERTopic & 24 & 1205/2400 & 50.2 \\
% MNLI & 24 & 1004/2143 & 46.9 \\
% \bottomrule
% \end{tabular}
% \label{tab:method_comparison}
% \end{table}

Cross-method LLM-as-a-judge evaluation results shown in Figure 1 reveal that independent automatic latent domain discovery methods converge on similar topic representations. K-means and BERTopic methods have an Adjusted Rand Index (ARI) of 0.42 and a Normalised Mutual Information (NMI) of 0.55, thus indicating moderate, positive agreement and predictability. Whereas the scores between MNLI and the others (K-means: ARI 0.08, NMI 0.26 and BERTopic: 0.07, 0.28) indicate that the entailment-based groupings are almost completely different to the alternative methods. Given that MNLI adapters are the only AutoAdapt method to win on RACE, MBPP, CRUXEval and HellaSwag, this supports the idea that the entailment-based domain discovery method not only identifies alternative, but also meaningful representations in comparison with the other AutoAdapt methods.

K-Means and BERTopic share representations relating to general knowledge, creative writing, finance and natural language inference. Although they both create three distinct medical clusters, only K-Means wins each of them. BERTopic's Medical Ethics cluster loses to single LoRA on LLM-as-a-judge methodology. BERTopic and MNLI overlap on creative writing and reasoning representations. Topic-level agreement breakdown indicates that MNLI creates a large adapter-losing domain for chain-of-thought reasoning that overlaps with most of BERTopic's representations. Overall, MNLI as an entailment method is not as uniform as K-Means and BERTopic.  

\subsubsection{Domain-Specific Benchmarks}

In addition to LLM-as-a-judge evaluation, domain-specific benchmarks shown in Table \ref{tab:cross_method_convergence} also reveal that AutoAdapt methodology achieves parity with single LoRA. For each benchmark we compute the accuracy difference between AutoAdapt and the single LoRA, with variance from the two binomial accuracies. We then combine these differences across the 14 benchmarks as an inverse-variance weighted mean, reporting 95\% CIs and $p$-values from a normal approximation. We find that K-means and MNLI methods are statistically indistinguishable from the single-LoRA baseline, while the BERTopic approach performs slightly worse. Each AutoAdapt configuration converges on GSM8K, Spider and MedMCQA wins. On the other hand, MultiNLI performance converging on a negative effect across each AutoAdapt approach indicates that a function applied to topically unconstrained content suffers from topic-based fragmentation. This effect is also observed across poor K-means and BERTopic AutoAdapt performance on BoolQ, RACE and HellaSwag. We hypothesise that enriching the topic embeddings with functional embeddings could help to produce clusters that group semantically and operationally similar samples together. We leave this as a promising direction for future work. 
\begin{figure}[t]
  % \centering
  % Defining both width and height forces the image to stretch and ignore its original aspect ratio
  \includegraphics[width=\columnwidth, height=11.5cm]{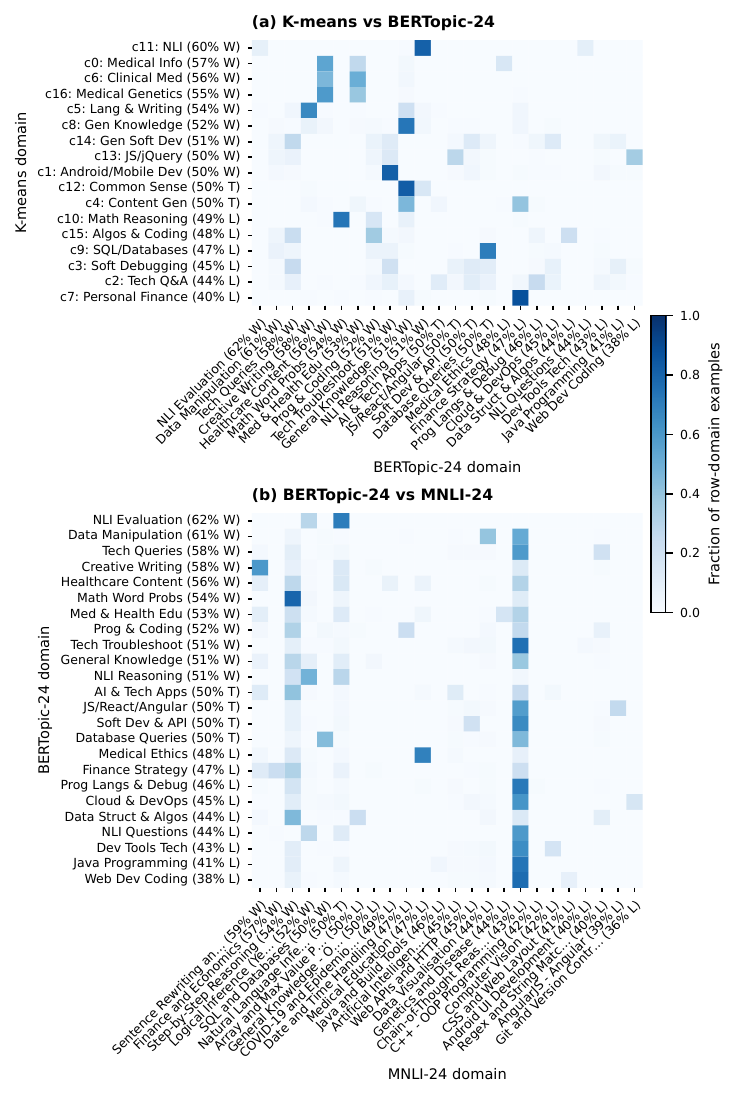}
  \caption{Domain overlap between discovery methods on LLM-as-a-judge evaluation with GPT-4o.}
  \label{fig:heatmapconvergence}
\end{figure}

\begin{table}[t]
\centering
\tiny
\caption{Cross-method benchmark deltas (percentage points vs.\ single LoRA). Sample size is $N=1000$ for all datasets except HumanEval ($N=164$), MBPP ($N=500$), and CRUXEval ($N=800$). K-means ($-0.27$pp, 95\% CI $[-1.36, +0.82]$, $p{=}0.63$) and MNLI ($+0.19$pp, 95\% CI $[-0.90, +1.28]$, $p{=}0.73$) methods do not differ significantly from single LoRA. BERTopic shows a small statistically significant negative aggregate effect ($-2.18$pp, 95\% CI $[-3.28, -1.08]$, $p{<}0.001$). CIs and $p$-values use an inverse-variance weighted mean of per-benchmark deltas.}
\begin{tabular}{lrrrl}
\toprule
\textbf{Benchmark} & \textbf{K-means} & \textbf{MNLI-24} & \textbf{BERTopic} & \textbf{Agreement} \\
\midrule
GSM8K     & +2.50  & +0.90  & +4.00  & WIN (all) \\
Spider    & +3.30  & +2.50  & +2.80  & WIN (all) \\
MedMCQA   & +0.70  & +5.20  & +4.00  & WIN (all) \\
BoolQ     & -0.60  & -1.60  & -2.50  & LOSE (all) \\
MultiNLI  & -7.80  & -8.50  & -13.70 & LOSE (all) \\
HumanEval & -7.32  & -4.27  & -4.27  & LOSE (all) \\
% \midrule
RACE      & -4.20  & +4.20  & -11.40 & mixed \\
PIQA      & +4.10  & +0.70  & -5.90  & mixed \\
Finance   & -0.30  & +4.90  & +0.50  & mixed \\
MATH      & +0.80  & +0.10  & -0.10  & mixed \\
PubMedQA  & +6.30  & -4.50  & +3.90  & mixed \\
MBPP      & -2.80  & +1.00  & -1.00  & mixed \\
CRUXEval  & -3.25  & +1.37  & -0.50  & mixed \\
HellaSwag & -2.60  & +2.20  & -7.90  & mixed \\
\bottomrule
\end{tabular}
\label{tab:cross_method_convergence}
\end{table}

Unlike with LLM-as-a-judge evaluation, the MNLI approach achieves the largest aggregate mean effect versus the baseline. MNLI performing relatively better on the benchmarks could suggest that it transfers better to out-of-distribution examples or adheres to the constrained output format required by datasets like RACE and HellaSwag, versus the open-ended generation that the held-out set addresses. Strikingly, coding adapters perform poorly across both benchmark and LLM-as-a-judge evaluation versus the single LoRA baseline. This consistent result suggests that diverse coding domains benefit from shared knowledge representation. Although MBPP, CRUXEval and HumanEval are topically similar, they assess different functions. HumanEval and MBPP are both synthesis tasks that require generating code from natural-language descriptions. Whereas, CRUXEval requires predicting the output for a given input. Including function-based embeddings in the automatic domain discovery process could improve coding performance by grouping operationally related samples together and therefore should be explored in future work. Supported by both domain-specific AutoAdapt and LLM-as-a-judge evaluation, we show that automatic domain specialisation is competitive with general-domain LoRAs, while offering modularity and extensibility that monolithic alternatives lack.

\subsection{Extensibility and Modularity}
\vspace{-0.39em}
Domain extensibility is a structural property of the AutoAdapt architecture. Existing domains are updated by retraining only the relevant adapter on new data, while new domains are incorporated by training a 
single additional adapter, in both cases leaving all other components unchanged. We simulate both extension scenarios: a domain update and a new domain addition. We report the results in Table \ref{tab:magicoder_heldout}.

Firstly, we simulate a post-deployment domain update by incorporating a 10,000-example subset of 
Magicoder~\cite{wei2024magicoderempoweringcodegeneration}. New samples are assigned domain labels via centroid routing. Nearly half (47\%) of the Magicoder samples are routed to the Algorithms \& Coding domain (c15). Including this new data results in minimal changes to the domain assignment of the 8,500-example test set, with 99.9\% of routing assignments remaining unchanged (8,494/8,500). The updated c15 adapter achieves the lowest negative-log-likelihood and perplexity scores across all configurations. Additionally, the GPT-4o judge evaluation indicates that the updated c15 adapter wins more than single LoRA across all configurations, despite single LoRA having access to 19× more training data.

Secondly, we simulate adding a new domain in a post-deployment scenario. A legal domain is added as it is not a distinct domain present in the original K-means domains. Using 1,000 legal instructions from LawInstruct to compute a new centroid, we re-route the original 8,500 test examples with 18 centroids. Routing stability is 99.7\% (8,472/8,500 unchanged). Manual inspection of the examples that changed to the legal domain indicates that many legal instructions did not have a dedicated domain before. Most examples (20/28) that changed domain relate to tax, liability and business-registration prompts. We evaluate the new adapter with a blind, reference-grounded pairwise GPT-4o judge on 200 held-out LawInstruct QA examples. Every query is routed to the legal adapter, and its response is compared with that of the original 17-domain AutoAdapt system and of the frozen single LoRA. The legal adapter response is preferred in 126/200 comparisons against the original 17-domain system and 129/200 against the single LoRA. Given that other existing adapters are not modified, previously learned domains are unaffected by construction, and no monolithic retraining is required.

In addition to extensibility, decomposing heterogeneous data into independent 
domains enables parallel adapter training. Each adapter trains on its own data partition without inter-adapter dependencies. In our experiments, 
three adapters are trained simultaneously on a 
single A100 80GB GPU, reducing wall-clock time 
by 1.7$\times$ relative to single LoRA (10h 8min 
vs 17h 0min) and 2.0$\times$ relative to full 
model fine-tuning (10h 8min vs 20h 22min). At 
inference time, parameter-free centroid routing 
adds relatively minimal latency in comparison with neural approaches, with the base model 
remaining resident in GPU memory while only the 
active lightweight adapter is swapped per request.
\begin{table}[t]
\centering
\tiny
\caption{Results of the post-deployment domain update scenario simulations. We report the average negative log-likelihood (NLL) and perplexity (PPL) as held-out language modelling metrics. We also report LLM-as-a-judge results with GPT-4o. }
\begin{tabular}{lrrrrr}
\toprule
& \multicolumn{2}{c}{\textbf{Held-out LM}} & \multicolumn{3}{c}{\textbf{GPT-4o judge}} \\
\cmidrule(lr){2-3} \cmidrule(lr){4-6}
\textbf{Model} & \textbf{NLL} $\downarrow$ & \textbf{PPL} $\downarrow$ & \textbf{Upd.\ wins} & \textbf{Other wins} & \textbf{Ties} \\
\midrule
Updated c15 (+ Magicoder)    & \textbf{0.1926} & \textbf{1.212} & -- & -- & -- \\
Old c15 (no Magicoder)       & 0.2805 & 1.324 & \textbf{104} & 60 & 36 \\
Single LoRA (frozen)         & 0.2813 & 1.325 & \textbf{119} & 50 & 30 \\
Single LoRA (retrained)      & 0.2024 & 1.224 & \textbf{85}  & 76 & 38 \\
Full FT (retrained)          & 0.4488 & 1.566 & \textbf{167} & 16 & 17 \\
% \bottomrule
\toprule
\textbf{Model} & \textbf{NLL} $\downarrow$ & \textbf{PPL} $\downarrow$ & \textbf{Legal wins} & \textbf{Other wins} & \textbf{Ties} \\
\midrule
Legal adapter (oracle)                    & 2.4817 & 11.961 & -- & -- & -- \\
AutoAdapt without Legal (17 domains)       & 2.3849 & 10.858 & \textbf{126} & 74  & 0 \\
Frozen single LoRA                         & \textbf{2.2971} & \textbf{9.945}  & \textbf{129} & 71  & 0 \\
% Base model (no adapter)                    & 2.9723 & 19.537 & 34  & 165 & 1 \\
\bottomrule
\end{tabular}
\label{tab:magicoder_heldout}
\end{table}

\section{Conclusion}

AutoAdapt establishes automatically discovered latent domain 
specialisation as a modular and practical alternative to 
monolithic instruction tuning. By adopting a taxonomy-free approach and parameter-free routing at 
inference, the framework matches monolithic single LoRA 
performance across domain-specific benchmarks and LLM-as-a-judge. We further demonstrate that domain extensibility is a structural property of AutoAdapt via post-deployment update simulations. Together, these findings indicate that AutoAdapt is a practical and extensible alternative to full retraining in heterogeneous and evolving deployment settings. Future work should extend beyond a low-resource deployment scenario and investigate the impact of AutoAdapt on larger models.

\begin{credits}
\subsubsection{\ackname} This publication has emanated from research conducted with the financial support of Taighde Éireann - Research Ireland under Grant number 18/CRT/6223.

\subsubsection{\discintname}
The authors have no competing interests to declare that are
relevant to the content of this article. 
\end{credits}
\bibliographystyle{splncs04}
\bibliography{custom}




% LNCS Homepage, \url{http://www.springer.com/lncs}, last accessed 2023/10/25
% \end{thebibliography}
\end{document}